%% file: main.tex
\documentclass[11pt,a4paper]{article}
\usepackage[hyperref]{conll-2019}
\usepackage{times}
\usepackage{latexsym}

\usepackage{url}
\usepackage{times}
\usepackage{subfiles}
\usepackage{latexsym}
\usepackage{graphicx}
\usepackage{amsmath}
\usepackage{amsfonts}
\usepackage{enumerate}
\usepackage{multirow}
\usepackage{float}
\usepackage{subfig}
\usepackage{enumitem}
\usepackage{subfig}
\usepackage{booktabs}
\graphicspath{ images/ }
 \usepackage{url}
\usepackage{todonotes}
\usepackage{enumitem}

\aclfinalcopy 

\title{Learning Analogy-Preserving Sentence Embeddings for Answer Selection}

\author{A{\"i}ssatou Diallo\textsuperscript{\dag,\ddag} , Markus Zopf\textsuperscript{\ddag} and  Johannes F{\"u}rnkranz\textsuperscript{\ddag} \\
 \textsuperscript{\dag}Research Training Group AIPHES\\
 \textsuperscript{\ddag}Knowledge Engineering Group \\
 Department of Computer Science, Technische Universit{\"a}t Darmstadt \\
 {\tt {\{diallo@aiphes, mzopf@ke, juffi@ke\}.tu-darmstadt.de}} }

\date{}

\begin{document}
\maketitle

\begin{abstract}
\hyphenpenalty=100
\subfile{abstract.tex}

\end{abstract}

\hyphenpenalty=1000
\section{Introduction}

\subfile{introduction.tex}

\section{Related Work}
\label{sec:related}
\subfile{related_work.tex}

\section{Analogical Embeddings}
\label{sec:embeddings}

\subfile{approach.tex}

\section{Experiment}
\label{sec:evaluation}

\subfile{experiment.tex}

\section{Conclusion}

This work introduced a new approach to learn sentence representations for answer selection, which preserve structural similarities in the form of analogies. Analogies can be seen as a way of injecting reasoning ability, and we express this by requiring common dissimilarities implied by analogies to be reflected in the learned feature space. We showed that explicitly constraining structural analogies in the learnt embeddings leads to better results over the distance-only embeddings.
We believe that it is worth-while to further explore the potential of analogical reasoning beyond their common use in word embeddings, as it is a natural mean of learning and generalizing about relations between entities. The focus of this work has been on answer selection, but analogical reasoning can be useful in many other machine learning tasks such as machine translation or visual question answering. As a next step, we plan to explore other forms of analogies that involve modelling across domains. 

\section*{Acknowledgments}
This work has been supported by the German Research Foundation as part of the Research Training
Group Adaptive Preparation of Information from Heterogeneous Sources (AIPHES) under grant No. GRK 1994/1.

\bibliography{conll-2019}
\bibliographystyle{acl_natbib}

\end{document}

%% file: abstract.tex

Answer selection aims at identifying the correct answer for a given question from a set of potentially correct answers.
Contrary to pre\-vious works, which typically focus on the semantic similarity between a question and its answer, 
our hypothesis is that question-answer pairs are often in analogical relation to each other.
Using analogical inference as our use case, we propose a framework and a neural network architecture for learning dedicated sentence embeddings that preserve analogical properties in the semantic space. We evaluate the proposed method on benchmark datasets for answer selection and demonstrate that our sentence embeddings indeed capture analogical properties better than conventional embeddings, and that analogy-based question answering outperforms a comparable similarity-based technique.   


%% file: introduction.tex

Answer selection is the task of identifying the correct answer to a question from a pool of candidate answers. The standard methodology is to prefer answers that are semantically similar to the question. Often, this similarity is strengthened by bridging the lexical gap between the text pairs via
learned semantic embeddings for words and sentences.
The main drawback of this method is that question-answer (QA) pairs are modeled 
independently, and that the correspondence between different pairs is not considered in these embeddings. In fact, 
these methods only focus on the relationship that may exist between the entities that constitutes the QA pair at hand and are thus, 
limited to pairwise semantic structures.

Instead, we argue in this paper that questions and their correct answers often form analogical relations. For example, the question ''Who is the president of the United States?'' and its answer are in the same relation to each other as the question ''Who is the current chancellor of Germany?'' and ''Angela Merkel''. Thus, for modelling these relations, we need to look at quadruples of textual items in the form of two question-answer pairs, and want to reinforce that they are in the same relation to each other. 

\begin{figure}[t]
    \centering
    \includegraphics[width=\columnwidth]{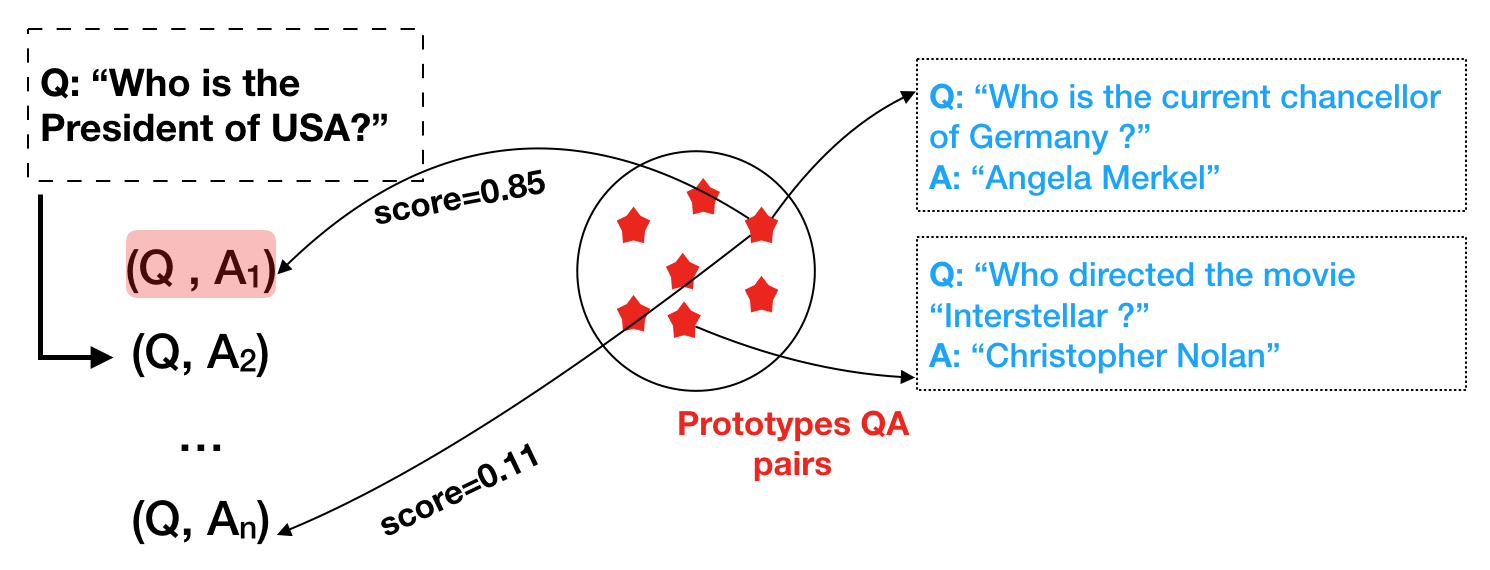}
    \caption{Illustration of analogy-based answer selection. Given a question and its candidate answers, each pair is compared to a QA prototype pair. The candidate answer with the highest score is assumed to be the correct answer.}
    \label{swer selection basedl reasoning.  Given someprototype QA pair, the goal identi   on afig:my_labelnalogica}
\end{figure}
We expect
that using analogies to identify and transfer positive relationships between QA pairs will be a better approach for tackling the task of answer selection than simply looking at the similarity between individual questions and their answers. 

We use sentence embeddings as the mechanism to assess the relationship between two sentences, and aim to learn a latent representation in which their analogical relation is explicitly enforced 
in the latent space. 
Analogies are defined as relational similarities between two pairs of entities, such that the relation that holds between the entities of the first pair, also holds for the second pair. Loosely speaking, the quadruple of sentences is in \emph{analogical proportion} if the difference between the first question and its answer is approximately the same as the difference between the second question and its answer.

This formulation is especially valuable because analogies allow to put on relation pairs that are not directly or explicitly linked. Consequently, in the vector space, analogous QA pairs will be oriented in the same direction, whereas dissimilar pairs will not correspond.

The remainder of the paper is organized as follows: the next section will present related work on answer selection, metric learning, as well as laying down the foundations of analogical reasoning. In Section~\ref{sec:embeddings}, we formally define analogies, and introduce our approach for 
learning such analogical embeddings. Finally, in Section~\ref{sec:evaluation}, we evaluate the learnt representations to demonstrate that the found embeddings indeed respect the sought analogies, and to illustrate the benefits of analogies for the task of answer selection.


%% file: related_work.tex
\paragraph{Answer Selection.} Answer selection is an important problem in natural language processing that has drawn a lot of attention in the research community \cite{lai2018review}. Given a question and a set of candidate answers, the task is to identify the correct answer(s) in this set. This task can be formulated as a classification or a ranking problem. Early works relied on computing a matching score between a question and its correct answer, and were characterized by the heavy reliance on feature engineering for representing the QA pairs.  Representative works include \cite{DBLP:conf/semeval/FiliceCMB16}, which studies the effects of various similarity, heuristic, and thread-based features, or \cite{DBLP:conf/cikm/TymoshenkoM15}, which analyzes the effect of syntactic and semantic features extracted by syntactic parser for answer re-ranking.
Recently, deep learning methods have achieved excellent results in mitigating the difficulty of feature engineering. These methods are used to learn latent representations for questions and answers independently, and a matching function is applied to give the score of the two texts. The most representative works in this line of work include \cite{DBLP:conf/acl/WangN15,yin2016abcnn,DBLP:conf/sigir/SeverynM15,DBLP:conf/sigir/TayPLH17}.

\paragraph{Embeddings and Metric Learning.} Our work is also related to representation learning usig deep neural networks. In fact, learning the embeddings of entities can be seen as a knowledge induction process, as those induced latent representations can be used to infer properties of unseen samples. 
Although many studies confirmed
that embeddings obtained from distributional similarity can be useful in a variety of different tasks,
\cite{DBLP:conf/naacl/LevyRBD15} showed that the semantic knowledge encoded by general-purpose similarity embeddings is limited, and that enforcing the learnt representations to distinguish functional similarity from relatedness is beneficial. For this purpose, many task-specific embeddings have been proposed for a variety of tasks including \cite{DBLP:conf/naacl/RiedelYMM13} for binary relation extraction and \cite{DBLP:conf/emnlp/FitzGeraldTG015} for semantic role labeling. This work aims to preserve more far reaching structures, namely analogies between pairs of entities.

\paragraph{Analogical Reasoning.} Analogical reasoning has been an active research topic in classic artificial intelligence. It has been successfully used in different domains such as classification \cite{DBLP:conf/ecai/BounhasPR14}, clustering \cite{DBLP:journals/jmlr/MarxDBS02}, dimensionality reduction \cite{DBLP:conf/nips/MemisevicH04}, or learning to rank \cite{DBLP:conf/aaai/FahandarH18}.  
\citet{DBLP:journals/cogsci/Gentner83} studies analogies with respect to human cognition, defines an analogy as a relational similarity over two pairs of entities, and differentiates it from the more superficial similarity defined by attributes. Since this general definition of analogy requires high-level reasoning which is not scalable to large-scale automated prediction systems, \citet{miclet2008analogical} define the concept of analogical dissimilarity between entities in the same semantic universe. The analogical dissimilarity allows to perform direct inference for unseen entities. Contrary to their direct inference setting, we enforce the analogical constraints in the learned embedding in the form of geometrical constraints, by imposing the co-linearity of the vector that maps the entities of each pair in the analogical proportion. It is worth mentioning that analogies have been found as the result of several word embedding models---inter alia \cite{DBLP:conf/nips/MikolovSCCD13,DBLP:conf/emnlp/PenningtonSM14}---but those are allegedly only empirical observations, which we found to not carry over to our task.


%% file: approach.tex

\newtheorem{definition}{Definition}

In this section, we explain our approach towards generating semantic embeddings that preserve analogical proportions.

\subsection{Analogical Reasoning}
\label{sec:analogical-reasoning}
In this section, we briefly introduce key concepts in analogical reasoning, starting with analogical proportions.
\begin{definition}[Analogical Proportion]
Let $a$, $b$, $c$, $d$ be four values from a domain $\mathbb{X}$. The quadruple $(a,b,c,d)$ is said to be in analogical proportion $a:b::c:d$ if $a$ is related to $b$ as $c$ is related to $d$, i.e., $\mathcal{R}(a,b) \sim \mathcal{R}(c,d)$.
\end{definition}
This comparative relation between two pairs of entities can be expressed in many ways \cite{dubois2016multiple}, but the most noteworthy are:
\begin{itemize}[noitemsep,nolistsep]
    \item[--] Arithmetic proportion: $ (a-b) = (c-d)$
    \item[--] Geometric proportion: $ \dfrac{\min(ad,bc)}{\max(ad,bc)} $
\end{itemize}
In this work, we focus solely on the arithmetic interpretation of analogy. 

\begin{figure}[b]
\centering
  \subfloat[]{\includegraphics[width=2cm]{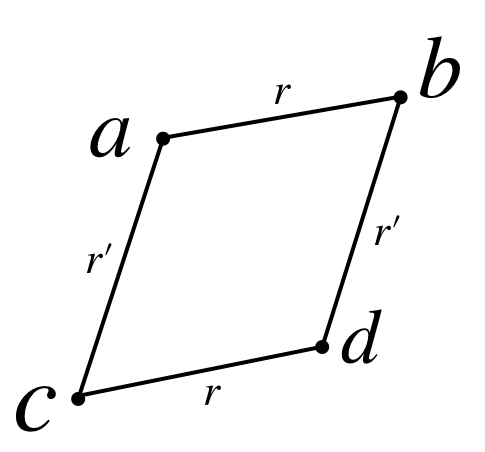} }%
    \qquad
    \subfloat[]{\includegraphics[width=2cm]{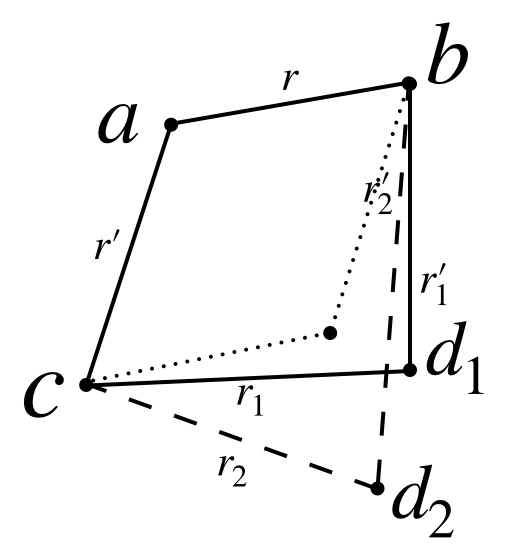} }%
    \caption{Analogical parallelograms in ${\mathbb{R}}^n$. (a) shows the case where $(a-b)=(c-d)$. The geometrical structure is a parallelogram. If $(a-b)\sim(c-d)$, the resulting structure is a general quadrangle with almost parallel sides (b). }%
    \label{fig:AD}%
\end{figure} 

An intuitive way of viewing analogies is through geometrical constraints in an Euclidean space. Enforcing the relational similarity between pairs of elements is equivalent to constraining the four elements to form a parallelogram. 

The left graph of Figure~\ref{fig:AD} illustrates such an 
\emph{analogical parallelogram}.
As we can see, in an \emph{analogical parallelogram}, there is not only a relation $\mathcal{R}$ holding between $(a,b)$ and $(c,d)$ respectively, but there must also hold a similar relation $\mathcal{R}'$ between $(a,c)$ and $(b,d)$.

We can now make a first step towards our problem, which is learning to identify correct answers according using analogical inference. Given the aforesaid quadruple, when one of the four elements is unknown, an analogical proportion becomes an analogical equation. 
\begin{definition}[Analogical Equation]
An \emph{analogical equation} has the form
\begin{equation}
    a:b::c:x
\end{equation}
where $x$ represents an unknown element that is in analogical proportion to $a$, $b$, $c$.
\end{definition}

In our setting, an exact solution to an analogical equation can often not be expected. Instead, we aim at 
finding the element $d_i$, among $n$ candidates, where the analogical proportion is as closely satisfied as possible. For example, in the right graph of Figure~\ref{fig:AD}, neither $d_1$ nor $d_2$ are perfect solutions to the analogical equation $a:b::c:x$, but $d_1$ seems to be a better solution than $d_2$.

In order to relax the equality constraint between the pairs of entities, and to generalize the formulation of analogical proportions beyond the Boolean case, \citet{miclet2008analogical} proposed to measure the 
degree of an analogical proportion using analogical dissimilarity.

\begin{definition}[Analogical Dissimilarity]
In a Euclidean space, the degree of analogical dissimilarity of a quadruple $(a,b,c,d)$ is defined as
\begin{equation}
    v(a,b,c,d) = \left\|(a-b) - (c-d)\right\|
\end{equation}
\end{definition}
This equation represents the relation $\mathcal{R}$ as the difference between the entities of the pair and $\sim$ as the difference between the previously the so expressed relation pairs. Obviously, $v(a,b,c,d) = 0$ if $(a,b,c,d)$ are in analogical proportion, and the value increases the less similar $(a-b)$ and $(c-d)$ are to each other.

This allows us
to re-frame the original problem of answer selection as a ranking problem, in which the goal is to select the candidate answer $d$ which minimizes the degree of analogical dissimilarity:
\begin{equation}
    d = \arg \min_{i} v(a,b,c,d_i)_{i=1,...,N}
\end{equation}

In the following sections, we will describe the details of the model by motivating the architectural choices.
\begin{table*}[ht!]
\begin{tabular}{c|l}
\hline
    \multicolumn{2}{c}{\textbf{"Where" questions}} \\
    \hline
    \texttt{Sentence A} & "Where was Abraham Lincoln born?"\\
    \texttt{Sentence B} & "On February 12, 1809, Abraham Lincoln was born Hardin County, Kentucky" \\
    \texttt{Sentence C} & "Where was Franz Kafka born?"\\
    \texttt{Sentence D} & "Franz Kafka was born on July 3, 1883 in Prague, Bohemia, now the Czech Republic."\\
\hline
 \multicolumn{2}{c}{\textbf{"Who" questions}} \\
 \hline
    \texttt{Sentence A} & "Who made the rotary engine automobile?"\\
    \texttt{Sentence B} & "Mazda continued work on developing the Wankel rotary engine." \\
    \texttt{Sentence C} & "Who discovered prions?"\\
    \texttt{Sentence D} & "Prusiner won Nobel prize last year for discovering prions"\\
 \hline
  \multicolumn{2}{c}{\textbf{"When" questions}} \\
 \hline
    \texttt{Sentence A} & "When was Leonardo da Vinci born?"\\
    \texttt{Sentence B} & "Leonardo da Vinci was actually born on 15 April 1452 [...] " \\
    \texttt{Sentence C} & "When did Mt St Helen last have significant eruption?"\\
    \texttt{Sentence D} & "Pinatubo's last eruption [...] as Mt St Helen's did when it erupted in 1980."\\
 \hline
\end{tabular}
\caption{Example of analogy between sentences. \texttt{Sentence A} and \texttt{Sentence B} constitutes the prototype QA pair in the analogical quadruples \texttt{Sentence C} and \texttt{Sentence D} are the QA pair at hand. }
\label{tab:example}
\end{table*}

\subsection{Generating Quadruples}
\label{sec:gen_quad}
In this work, we consider QA pairs as relational data. We aim to transfer knowledge from pairs whose relation is well known, which we call \emph{prototypes}, to unseen pairs. For this, we train a model to encode analogies in the latent representations of the sentences. For creating a instances of quadruples to train the model, we adapt state-of-the-art datasets. 

An analogy quadruple has the following form: 
\begin{equation*}
    [q_p:a_p::q_i:a_{ij}]
\end{equation*}
where the $q_p$ and $a_p$, respectively stand for the question and the answer of the prototype pair, whereas $q_i$ is the $i$-th question and $a_{ij}$ is the $j$-{th} candidate answer to $q_i$. 

\begin{figure}[ht]
    \centering
    \includegraphics[width=0.8\columnwidth]{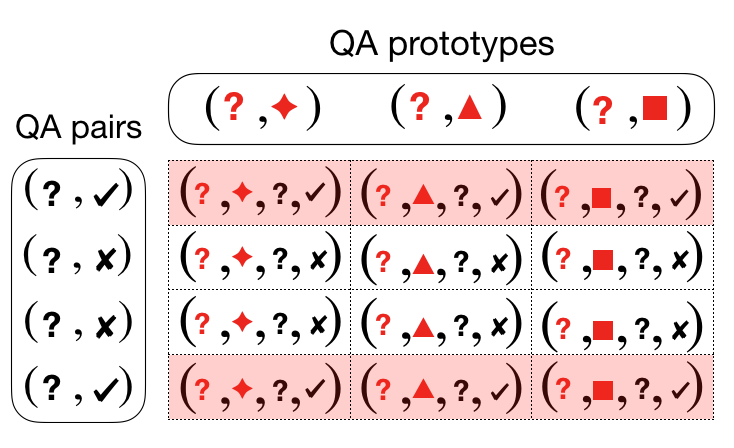}
    \caption{Procedure to generate analogical quadruples. The cells in red represent positive analogical quadruples, composed of a prototype QA pair, a question and its correct answer. In reverse, a negative quadruple contains a QA prototype, a question and one of its incorrect answer.}
    \label{fig:gen_quad}
\end{figure}

Given a set of questions and their relative candidates answers, we construct the analogical quadruples in two steps. First, we divide all the questions into three different subsets of wh-word questions: \textit{"Who"}, \textit{"When"} and \textit{"Where"}. We focus on these three types because their answer type fall in distinct and easily identifiable categories:
\begin{itemize}
    \item "Where" corresponds to an answer of type "Location"
    \item "Who" corresponds to an answer of type "Person"
    \item "When" corresponds to an answer of type "Date" or "Time"
\end{itemize}

Table~\ref{tab:example} illustrates examples of quadruples for the three described categories.
From these categories, we extract a variable number of QA pairs in order to form the prototype set. To generate positive quadruples, we select a prototype from one of the above-mentioned subsets and we associate a question from the same set and the correct answer among its candidates. This procedure provides a large number of analogical quadruples. 
On the other hand, to generate negative training samples we use the following approach: in the same subset, we associate a prototype, a question and a randomly selected wrong answer among its candidates. This is done in order to purposely break the analogical relation between a prototype QA pair and the QA pair at hand. This approach will generate a set of hard examples to help improve the training. Figure~\ref{fig:gen_quad} illustrates the procedure. 

To summarise, ranking by analogical dissimilarity is performed in three steps:
\begin{enumerate}
    \item Given a prototype QA pair, a question and $N$ candidate answers, $N$ quadruples are generated.
    \item The analogical dissimilarity score is computed for each quadruple.
    \item The $N$ candidates are consequently ranked by the analogical dissimilarity score.
\end{enumerate}
The next section closely describe the architectural choices of the model.
\subsection{Quadruple Siamese Network}

\begin{figure}[ht]
    \includegraphics[width=50mm]{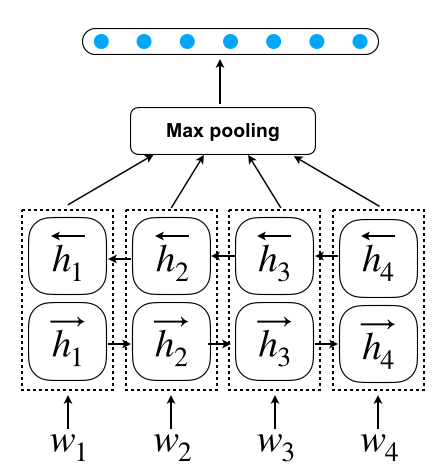}
    \centering
     \caption{BiGRU with max pooling.}
    \label{fig:maxpool}
\end{figure}

We recall that our focus is on learning an embedding function that pushes analogous QA pairs with similar mappings to be mutually close by enforcing a geometrical constraint in the vector space. This constraint states that the vector shift that maps the entities of the first pair should be similar to the vector shift of the second pair, according to the degree of analogical dissimilarity that holds between the two pairs $(a,b), (c,d)$. 

\begin{figure*}[ht]
    \includegraphics[width=0.85\textwidth]{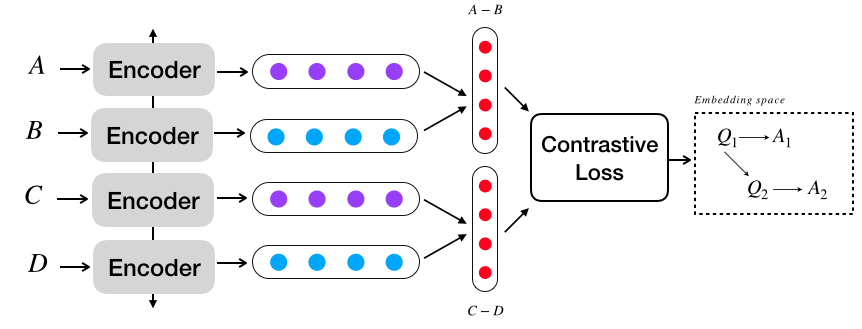}
    \centering
     \caption{Siamese architecture.}
    \label{fig:architectures}
\end{figure*}

To tackle this problem, we propose a Siamese network architecture as shown in Figure~\ref{fig:architectures}. In the next paragraphs, we describe the notation used and the details of each component of the model. 

\paragraph{Notation.} Let $\mathcal{Q}$ and $\mathcal{A}$ be the space of all questions and candidate answers. We denote a quadruple of sentences as $(a,b,c,d)$, where $a,c \in \mathcal{Q}$ and $b,d \in \mathcal{A}$. Quadruples are assigned a label $y = 1$ if the analogical proportion holds, and $0$ otherwise. $\theta$ denotes the parameters to be learnt that map the relation from $a$ to $b$, and $c$ to $d$ respectively. Let $x_{\cdot}$ refer to the latent representation of one sentence in the quadruple. 

\paragraph{Architecture.} The Siamese network takes as input four sentences. The sub-networks in the Siamese model share the parameters and learn the vector representations for every sentence received as input. A sentence $S_i = {w_{i1},...w_{ik}}$ where $w_{ij}$ represents the $j^{th}$ word in the sentence $S_i$, $\forall i \in 1 \leq i \leq n$ and $\forall j \in 1 \leq j \leq k$. 
Words are mapped into word embeddings $x_{ij}=Ew_{ij}$, where $E^{d,|V|}$ is a matrix of vectors of size $d$, and $V$ is the vocabulary. Out-of-vocabulary words are initialized by a random vector. 
We use bidirectional gated recurrent units (GRUs) \cite{DBLP:conf/ssst/ChoMBB14} over the input sentence. For a sentence of $T$ words, the network encodes $T$ hidden states $h_1,...,h_T$ such that:
\begin{equation*}
\begin{gathered}
  \overrightarrow{h_t}= \overrightarrow{GRU}_t(w_1,...,w_T)\\
   \overleftarrow{h_t}= \overleftarrow{GRU}_t(w_1,...,w_T)\\
   h_t = [\overrightarrow{h_t},\overleftarrow{h_t}]
\end{gathered}
\end{equation*}
In order to obtain a fixed-size vector, we select the maximum value over each dimension of ${h_t}$ using max pooling. 
After this step, we obtain four vectors of dimension $d$, one for each input sentence of the quadruple. 

\paragraph{Training Strategy.} 
The next step is to get the semantic relation between the pairs of input sentences. Given a pair a vectors, $(x_i,x_j)$, the arithmetic proportion expects the difference of the vectors to encode the relational similarity between the entities that constitutes the pair.
We let the network predict four $d$-dimensional embedding vectors, which we merge through a pairwise subtraction. Let $f_W(\cdot)$ be the projection of an input sentence in the embedding space computed by the network function $f_W$.
 Furthermore, let

\begin{eqnarray}
    f_{ab} = f_W(a) - f_W(b) \\
    f_{cd} = f_W(c) - f_W(d)
\end{eqnarray}

be the pairwise differences between the embedding vectors.
In order to separate instances of analogical proportion, similar pairs need to be mapped mutually close to each other, whereas dissimilar instances should be pushed apart.

For the energy of the model, we use the cosine similarity between the vector shifts of each pair of the quadruple:
\begin{equation}
    E_W(f_{ab}, f_{cd}) = \frac{f_{ab}\cdot f_{cd}} {\|f_{ab}\|\|f_{cd} \|}
    \label{eq:energy}
\end{equation}
We argue that this is an appropriate energy function  since the goal is for the pairs of parallel vectors to be parallel which maximises the analogical parallelogram likelihood.

We propose to use the contrastive loss \cite{hadsell2006dimensionality}
to perform the metric learning. This loss function has two terms, one for the similar and and another dissimilar samples. The similar instances are denoted by a label $y=1$ whereas the dissimilar pairs are represented by $y=0$. Thus, the loss function has the following form:
\begin{equation}
    \mathcal{L}_W = y\  \mathcal{L}_+(f_{ab}, f_{cd})+\ (1-y) \mathcal{L}_-(f_{ab}, f_{cd})
\end{equation}
Each term is expressed by:
\begin{eqnarray}
 \mathcal{L}_+(f_{ab}, f_{cd}) = (1-E_W)^2 \\
    \mathcal{L}_-(f_{ab}, f_{cd}) = \max((E_W-m)^2,0)  
\end{eqnarray}

This loss function measures how well the model learns to encode similar transformations such that analogous pairs are mutually close and form an analogical parallelogram in the embedding space, while pushing dissimilar transformations apart.
Given a question and a pool of candidate answers, the goal is to rank the correct answer in the first position, based on how well each sentence completes the analogical equation according to~\eqref{eq:energy}.

This architecture is summarized in Figure \ref{fig:architectures}. 
We learn all the parameters of the model through a gradient based method that minimizes the L2-regularized loss. Further details about the implementation are given in section \ref{sec:experiment}. 


%% file: experiment.tex

In this section, we present an evaluation of our approach in two experiments: first, in Section~\ref{sec:eval-analogy}, we confirm that 
the found analogical embeddings do indeed improve 
the analogical parallelogram structure illustrated in Figure~\ref{fig:AD} over commonly used word- and sentence-based embeddings. In Section~\ref{sec:eval-QA} we then show that this also results in improved performance for question answering. Before that, we start with a brief description of our experimental setup.

\subsection {Experimental Setup}
\label{sec:experiment}

We begin the assessment of our model with a direct evaluation, which is ranking candidate answers in the same setting as during the training of the embedding. We generate quadruples with the same prototypes used for the training and we look for the correct answers by iteratively solving the analogical equations. We compare our model to commonly used sentence representations methods to evaluate the proposed approach results with respect to general purpose sentence embedding and word embedding methods.
In the next paragraphs we present the experimental setup and the results obtained.

\paragraph{Datasets.} We validate the proposed method on two datasets: WikiQA \cite{DBLP:conf/emnlp/YangYM15}, an open domain QA dataset with answers collected over Wikipedia and TrecQA, which was created from the TREC Question Answer Track. Both resources are well established for benchmarking answer selection. We split each dataset into three subsets, which contain only "who", "where" and "when" questions.  Table \ref{tab:data_by_question} reports the statistics of the two datasets.

\begin{table}[ht!]
\centering
\begin{tabular}{|c|c|c|c|c|c|c|c|}
\hline
    \multirow{2}{*}{}& \multicolumn{3}{c|} {WikiQA} &   \multicolumn{3}{c|} {TrecQA}\\
    \hline
    type & train & dev & test & train & dev & test \\
    \hline
    "Who" & 119 & 15 & 34 & 190 & 11 & 8 \\
    "When" & 86 & 11 & 16 & 116 & 13 & 19 \\
    "Where" & 71 & 17 & 22 & 96 &  9 & 11 \\
    \hline
    Comb. & 276 & 43 & 72 & 402 & 33 & 38 \\
    \hline
\end{tabular}
\caption{Dataset by question type.}
\label{tab:data_by_question}
\end{table}

\paragraph{Evaluation metrics.} We assess the performance of our method by measuring the Mean Average Precision (MAP) and the Mean Reciprocal Rank (MRR) for the generated quadruples in the test set. Given a set of questions $Q$, MRR is computed as follows:
\begin{equation}
    \text{MRR} = \frac{1}{|Q|}\sum_{i=1}^{|Q|}\frac{1}{rank_i}
\end{equation}
where $rank_i$ represents the rank position of the first correct candidate answer for the $i^{th}$ question. In other words, MRR is the average of the reciprocal ranks of results for the questions in set $Q$. 

MAP is calculated as follows:
\begin{equation}
    \text{MAP} = \frac{1}{|Q|}\sum_{j=1}^{|Q|}\frac{1}{m_j}\sum_{k=1}^{|m_j|}\text{Precision}(\pi_{jk})
\end{equation}
where $q_j \in Q$ is a question whose candidate answers are ${a_1, ..., a_{m_{j}}}$ and $\pi_{jk}$ is the rank associated with those candidate answers. While MRR measures the rank of any correct answer, MAP computes the rank of all correct answers. Generally, MRR is higher than MAP on the same set of ranked objects.

\paragraph{Implementation details.} We initiate the embedding layer with FastText vectors. These weights are not updated during training. The dimension of the output of the sentence encoder is $300$. For alleviating overfitting we apply a dropout rate of $0.5$. The model is trained with Adam optimizer with a learning rate of $0.001$ and a weight decay rate of $0.01$.

\subsection{Quality of Analogical Embedding}
\label{sec:eval-analogy}

\paragraph{Baselines.} To support our claim that the learnt representations of our model encode the semantic of question answer pairs better than pre-trained sentence representation models, we choose four baselines commonly used to encode sentences: 

\begin{enumerate}
    \item  \textit{Word2Vec and Glove} \cite{DBLP:conf/nips/MikolovSCCD13, DBLP:conf/emnlp/PenningtonSM14}: We use the simple approach of averaging the word vectors for all words in a sentence. This method has the drawback of ignoring the order of the words of the sentence, but has shown to perform reasonably well. 
    \item \textit{InferSent} \cite{DBLP:conf/emnlp/ConneauKSBB17}: Sentence embeddings obtained from training on Natural Language Inference dataset. 
    \item \textit{Sent2Vec} \cite{pagliardini2017unsupervised}: A method to learn sentence embeddings such that the average of all words and n-grams can serve as sentence vector.
\end{enumerate}

For each document in test set, we generate analogical quadruples as explained in section \ref{sec:gen_quad}. Given a question $q_i$ in the test set with $k$ candidate answers, we obtain $p \times k$ possible quadruples, where $p$ is the cardinality of the prototype set. The network encodes each sentence in the quadruple and computes the cosine similarity \eqref{eq:energy} between the obtained vector shifts.

Not every prototype QA pair will fit to the QA pair at hand, so we compute $p \times m$ scores, and choose only the prototype that leads to the highest analogical score for each document and discard the other comparisons. One might think about using the average of the scores and sorting the candidate answers accordingly, but this strategy introduces noise in the analogical inference procedure.

\paragraph{Results.}
We applied the described procedure to vectors obtained from our network as well as from the baseline representation methods. The results are shown in Table~\ref{tab:quadruples}. 

\begin{table}[b!]
\centering
\begin{tabular}{c|c|c|c|}
\cline{2-4}
    & \multirow{2}{*}{Model} & \multicolumn{2}{c|} {WikiQA} \\
    \cline{3-4}
& & MAP & MRR \\

\hline
   \multicolumn{1}{|c|}{\multirow{2}{*}{W.E}} & Glove & 0.464 & 0.475  \\
    \multicolumn{1}{|c|}{} & Word2Vec & 0.432 & 0.453 \\
    \hline
    \multicolumn{1}{|c|}{\multirow{2}{*}{S.E}} & InferSent & 0.399 & 0.404  \\
   \multicolumn{1}{|c|}{}& Sent2Vec & 0.481 & 0.486 \\
    \hline
    & \textbf{This work} & \textbf{0.677} & \textbf{0.684}  \\
    \cline{2-4}
\end{tabular}
\caption{Evaluation on quadruples. W.E. indicates averaging over word embeddings approach. S.E indicates dedicated sentence embeddings.}
\label{tab:quadruples}
\end{table}

In order to better perceive the analogical properties of the baselines and the proposed approach, we also include a random baseline in the comparison.
We observe that averaging word embeddings such as Glove or Word2Vec performs better than the dedicated sentence representations in the \mbox{WikiQA} dataset.  This might be due to the fact that word embeddings have shown to encode some analogical properties. On the other hand, sentence embeddings have been trained with a particular learning objective, for example, InferSent has been train for the task of claim entailment with a classification objective and might not be suitable for representing relations between pairs of sentences. Nevertheless, ranking by the cosine similarity of the difference vectors do not lead to acceptable performances. This confirms our hypothesis that pre-trained sentence representation do not preserve analogical properties.

\begin{figure}[t]
    \centering
    \includegraphics[width=0.8\columnwidth]{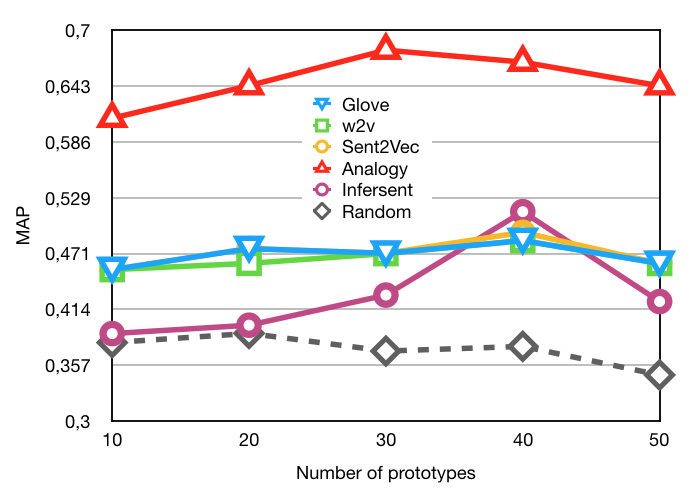}
    \caption{MAP for different number of prototypes.}
    \label{fig:map_proto}
\end{figure}

\begin{figure}[t]
    \centering
    \includegraphics[width=0.8\columnwidth]{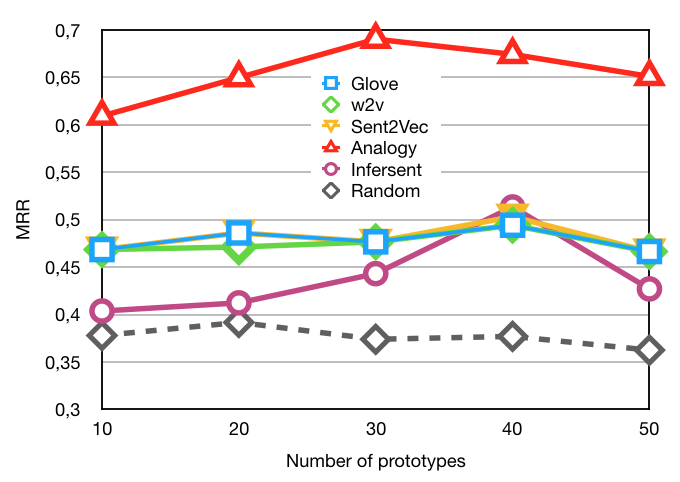}
    \caption{MRR for different number of prototypes.}
    \label{fig:mrr_proto}
\end{figure}

Similarly, we measure the influence of the number of prototypes on the performances. 
We vary the number of prototypes pair $p \in \{10,20,30,40,50\}$ and measure the MAP and the MRR for both datasets. The results are shown in Figures~\ref{fig:map_proto} and~\ref{fig:mrr_proto}. We can observe that the best performances are obtained for $p=30$ and that after both MAP and MRR decrease. The reason might be that a high number of prototypes brings more comparisons and increases the probability of spurious interactions between QA prototypes and QA pairs.

\subsection{Question-Answering Performance}
\label{sec:eval-QA}

A natural benchmark model for our work is 
the approach of \cite{DBLP:conf/ijcai/TamHNM17}, which is similar to ours in that it proposed to replace wh-word in questions with appropriate named entities. This approach leverages typological information from a named entity recognizer and the word vector space.
 
It showed that simply replacing the wh-word, with a named entity that has the highest cosine similarity with all the candidate answers for a given question. This substitution is operated for "where", "when" and "who" types of questions. Finally, the transformed QA pair is fed to a network suited for the task of answer selection. This study demonstrated that this simple pre-processing step improves the state of the art results for the task of answer selection. 

Alike our experimental setup, they divide the dataset in three categories, namely "when", "who" and "where", which is the same division we used for our experimental setup, and evaluate their method on the split dataset and the full dataset. 
We will consider their work as our baseline in order to evaluate the capabilities of the analogy based embeddings. 
Moreover, we compare our approach to a setup which doesn't exploit analogical properties. This is to say, a Siamese network that takes as input a question and a candidate answer, generate the respective representations and compute the cosine similarity of the obtained sentence embeddings. The described baseline corresponds to the model proposed by \cite{tan2015lstm} except for the fact that we use BiGRU for fair comparison.

The results are shown in Tables~\ref{tab:mrr_wikiqa} and~\ref{tab:mrr_trecqa}. 

We observe that simply computing the cosine similarity between the difference vector of the prototype QA pair and the QA pair at hand with the learnt embedding from the proposed approach lead to significant improvements for some particular type of questions. The bold numbers in Tables \ref{tab:mrr_wikiqa} and \ref{tab:mrr_trecqa} indicate the best results for each dataset. We can see that our method improves the MRR of at least two of questions types by a relevant margin. The last row of the same tables confirms that enforcing analogical properties in the embedding space generally improves the overall MRR for these three subsets. 

\begin{table}[t!]
\centering
\begin{tabular}{|c|c|c|c|}
\hline
    \multirow{2}{*}{}& \multicolumn{3}{c|} {WikiQA} \\
    \cline{2-4}
& Baseline & Tam et al. & Analogy \\

   \hline\hline
   "Who"  & 0.663 & 0.702 & \textbf{0.763} \\
   "When" & 0.582 & 0.664 & \textbf{0.701} \\
   "Where" & 0.568 & \textbf{0.616} & 0.602 \\
   \hline
    \textbf{Comb.} & 0.609 & 0.678 & \textbf{0.684} \\
   \hline
\end{tabular}
\caption{MRR on WikiQA.}
\label{tab:mrr_wikiqa}
\end{table}

\begin{table}[t!]
\centering
\begin{tabular}{|c|c|c|c|}
\hline
    \multirow{2}{*}{}& \multicolumn{3}{c|} {TrecQA} \\
    \cline{2-4}
& Baseline & Tam et al. & Analogy \\

   \hline\hline
   "Who" & 0.787 & 0.781 & \textbf{0.981} \\
   "When" & 0.797 & \textbf{0.921} & 0.863 \\
   "Where" & 0.894 & 0.864 & \textbf{0.929} \\
   \hline
    \textbf{Comb.} & 0.837 & 0.875 & \textbf{0.909} \\
   \hline
\end{tabular}
\caption{MRR on TrecQA.}
\label{tab:mrr_trecqa}
\end{table}
